\documentclass[conference]{IEEEtran}
\IEEEoverridecommandlockouts
\usepackage{amsmath,amssymb,amsfonts}
\usepackage{algorithmic}
\usepackage{graphicx}
\usepackage{textcomp}
\usepackage{xcolor}
\usepackage{graphicx}
\usepackage{amsmath}
\usepackage{cite}
\usepackage{wrapfig}
\usepackage{soul}
\usepackage{booktabs}
\usepackage[colorlinks,urlcolor=blue,linkcolor=blue,citecolor=blue]{hyperref}
\usepackage{mdframed}
\usepackage{tcolorbox}
\usepackage{fancyhdr}

\fancypagestyle{firststyle}
{
   \fancyhf{}
    \lhead{\small Cite as: \textit{R. Wickramarachchi, C. Henson and A. Sheth, "Knowledge-Based Entity Prediction for Improved Machine Perception in Autonomous Systems," in IEEE Intelligent Systems, 2022, doi: 10.1109/MIS.2022.3181015}.}
}

\def\BibTeX{{\rm B\kern-.05em{\sc i\kern-.025em b}\kern-.08em
    T\kern-.1667em\lower.7ex\hbox{E}\kern-.125emX}}
\begin{document}

%

\title{Knowledge-based Entity Prediction for Improved Machine Perception in Autonomous Systems}

\author{\IEEEauthorblockN{\textbf{Ruwan Wickramarachchi}}
\IEEEauthorblockA{Artificial Intelligence Institute\\
University of South Carolina\\
Columbia, SC, USA \\
\texttt{ruwan@email.sc.edu}}
\and
\IEEEauthorblockN{\textbf{Cory Henson}}
\IEEEauthorblockA{Bosch Research and Technology Center \\
Pittsburgh, PA, USA \\
\texttt{cory.henson@us.bosch.com}}
\and
\IEEEauthorblockN{\textbf{Amit Sheth}}
\IEEEauthorblockA{Artificial Intelligence Institute\\
University of South Carolina\\
Columbia, SC, USA \\
\texttt{amit@sc.edu}}
}

\maketitle
\thispagestyle{firststyle}
\begin{abstract}
Knowledge-based entity prediction (KEP) is a novel task that aims to improve machine perception in autonomous systems. KEP leverages relational knowledge from heterogeneous sources in predicting potentially unrecognized entities. In this paper, we provide a formal definition of KEP as a knowledge completion task. Three potential solutions are then introduced, which employ several machine learning and data mining techniques. Finally, the applicability of KEP is demonstrated on two autonomous systems from different domains; namely, autonomous driving and smart manufacturing. We argue that in complex real-world systems, the use of KEP would significantly improve machine perception while pushing the current technology one step closer to achieving full autonomy.\\
\end{abstract}

\begin{IEEEkeywords}
entity prediction, machine perception, autonomous driving, smart manufacturing, event perception, knowledge-infused learning
\end{IEEEkeywords}

\section{Introduction}
\label{sec:intro}
Autonomous systems are becoming an integrated part of everyday life with applications in various domains, including transportation, manufacturing, and healthcare. The primary goal of autonomous systems is to imbue machines with intelligence that enables them to sense and learn to function within changing environments. Achieving this goal, however, entails solving many hard AI problems. For example, autonomous driving (AD), a popular application area in autonomous systems, often functions as a test-bed for tackling such complex AI problems. Scene understanding is a particularly important challenge in machine perception that involves sense-making at different levels; including visual scene detection, recognition, localization, causal reasoning, etc. \cite{ramanishka2018toward}. In this paper, we define the problem of knowledge-based entity prediction (KEP), a novel formulation that aims to improve current scene understanding technology by leveraging structured relational knowledge.\\

\begin{tcolorbox}[title=KEP Definition]
\textit{KEP is the task of predicting the inclusion of potentially unrecognized entities in a scene, given the current and background knowledge of the scene represented as \textbf{a knowledge graph.}}
\end{tcolorbox}

\begin{wrapfigure}{r}{0.32\textwidth}
    \centering
    \includegraphics[width=0.32\textwidth]{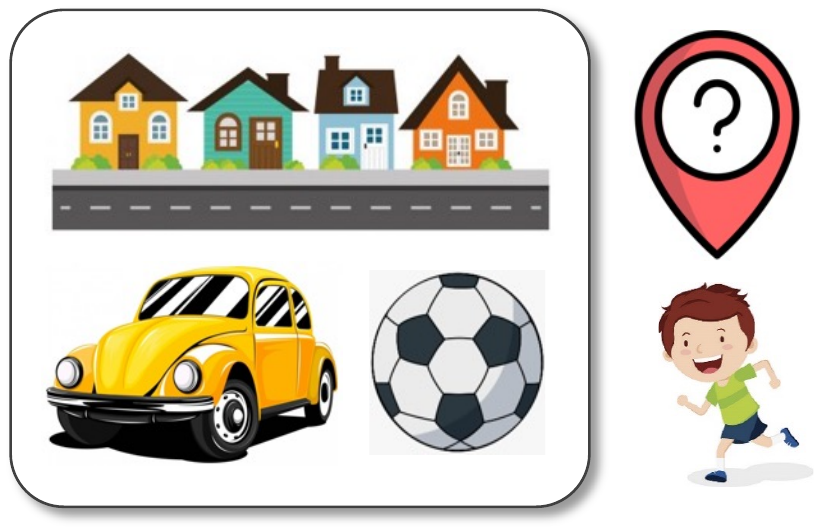}
    \caption{Knowledge-based Entity Prediction Problem in AD}
    \label{fig:kep}
\end{wrapfigure}

There are several reasons why a scene may contain unrecognized entities, including (but not limited to): sensor failure, occlusion, low resolution (e.g., from weather), or errors in computer vision models for object detection and recognition. As an example, consider this real-world scenario for autonomous driving: An autonomous vehicle is driving through a residential neighborhood and a ball is detected bouncing on the road (see Figure \ref{fig:kep}). KEP may be used to predict the likelihood of encountering a child chasing after the ball, even though no child has yet been detected.

Making such predictions, however, requires a deep understanding of the various aspects of the scene, including the semantic relations among the detected entities. For example, the knowledge that the bouncing ball is a type of toy and children often play with toys and that children often live and play in residential neighborhoods. Note that this type of semantic background knowledge is out-of-scope for current perception systems where computer vision plays the predominant role in assigning semantic types to the objects detected (i.e. semantic segmentation, object detection, and recognition) \cite{grigorescu2020survey}. \\

\begin{tcolorbox}[title=Hypothesis]
We hypothesize that potentially unrecognized entities in a scene may be detected through (1) the use of a knowledge graph to provide an expressive, holistic representation of scene knowledge, and (2) the application of knowledge-infused learning for knowledge completion; i.e. predicting missing knowledge in the graph.
\end{tcolorbox}

Contributions of this paper include: 
\begin{enumerate}
    \item Introducing the novel proposition of using KEP for improving machine perception in autonomous systems.
    \item Defining and formalizing KEP as a knowledge completion problem.
\end{enumerate}

The rest of the paper is structured as follows. Section \ref{sec:perception} discusses the proposition of using KEP to improve the machine perception in autonomous systems, along with its connection to perception in Psychology. Section \ref{sec:formalization} formalizes KEP as a knowledge completion problem and introduces three potential solutions. Section \ref{sec:case-studies} presents two case studies for using KEP in different application domains, autonomous driving and smart manufacturing, while Section \ref{sec:results} presents experimental results of a KEP implementation on AD data. Finally, we wrap up in Section \ref{sec:conclusion} with conclusions and future work.

\section{Improved Machine Perception Through KEP} 
\label{sec:perception}

\begin{figure*}[ht]
    \centering
    \includegraphics[width=1.0\textwidth]{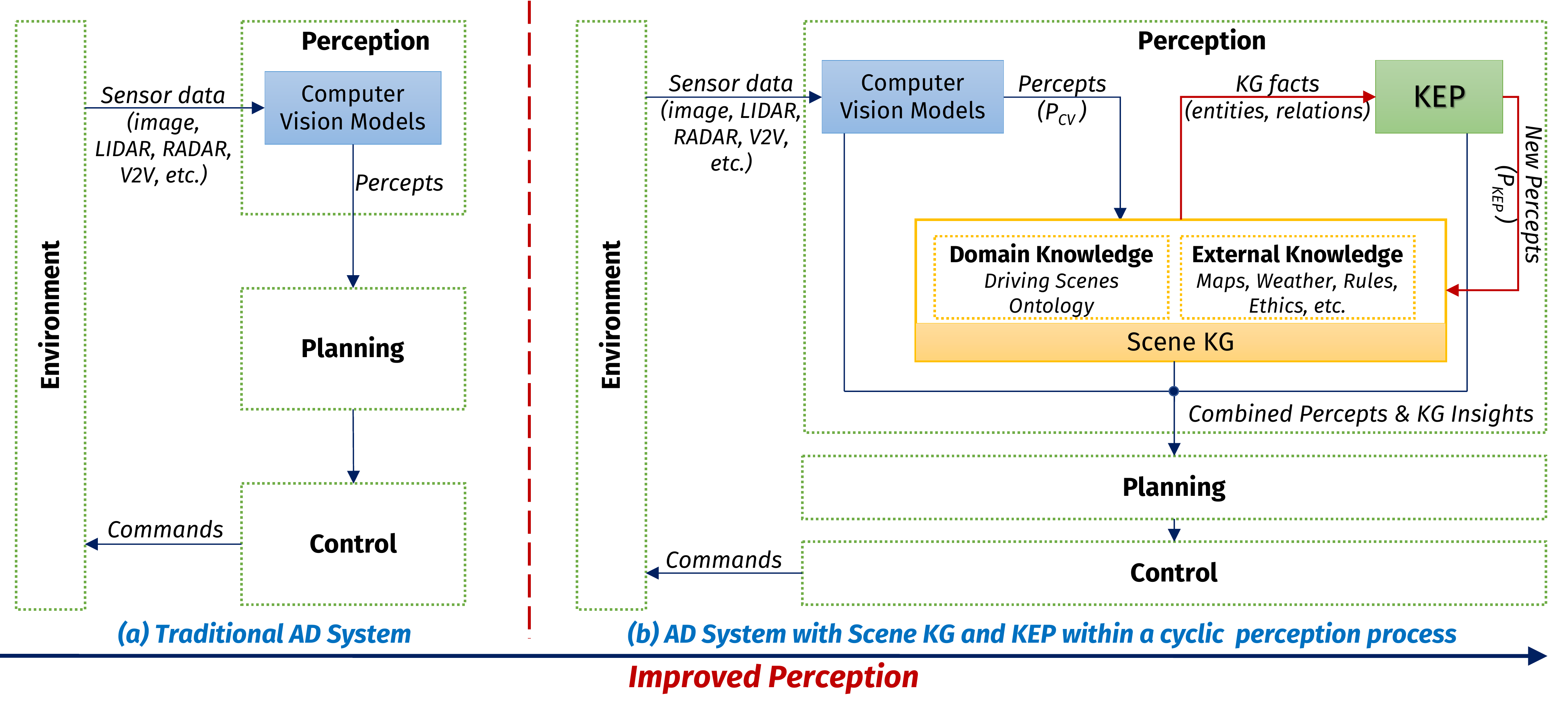}
    \caption{Improved machine perception through KEP in Autonomous Driving }
    \label{fig:ad_system}
\end{figure*}

The stated hypothesis allows us to view scene understanding in autonomous systems from a new perspective. For example, consider the following case from the autonomous driving domain. The perception module, one of the three main functional components in an AD system, uses the raw data from sensors to perform two critical tasks: (i) environmental perception -- obtaining percepts (i.e. detecting and recognizing high-level entities) in the environment ($P_{CV}$)), and (ii) localization -- determining the relative position of the ego vehicle and other entities in the environment. The outputs of the perception module are then used by subsequent processes (see Figure \ref{fig:ad_system}(a)) such as planning and control. The planning module performs tasks such as behavioral and motion planning of the ego vehicle while the control module governs the actuators that change the vehicle state and trajectory. This current process, however, has a serious vulnerability. If there is a mistake in the perception process, an unrecognized entity for example, this mistake will propagate through the subsequent processes and change the desired outcome. In the case of autonomous driving, this could lead to a fatal accident. \\

\textbf{Proposition.} With this background, consider the following proposition for improved machine perception. Environmental percepts of scenes (recognized by computer vision algorithms) ($P_{CV}$) are now represented within a KG. The KEP process then predicts a new set of percepts ($P_{KEP}$) for the potentially unrecognized entities -- i.e. unrecognized objects and events. These percepts are used to enrich the scene KG by adding new nodes and relations. Note that, as shown in Figure \ref{fig:ad_system}(b), this is a \textit{cyclical process} in which the scene KG is updated with new facts continuously over time. From a scene understanding perspective, the perception module's understanding of a scene improves as the KG becomes more complete. As a result of this process, (i) the perception module is enriched with potentially missing entities and (ii) the subsequent processes (i.e. planning and control) will have the benefit of using a richer set of percepts about the environment $P_{CV} + P_{KEP}$ along with other contextual and background cues from the relational knowledge in the scene KG. For example, consider the case where the perception module detects a pedestrian ($P_{CV}$) on the road. It does not, however, recognize that the pedestrian is \texttt{\textit{jaywalking}}. Even if no \texttt{\textit{jaywalking}} events have been seen while training the CV perception module, representing knowledge of this event -- i.e. \texttt{\textit{(Pedestrian, participatesIn, Jaywalking)}} -- in the scene KG could provide a new insight or cue for handling this \textit{edge-case} with KEP (i.e. $P_{KEP}$). We believe KEP has the potential to improve the machine perception in autonomous systems by significantly reducing the omission and misclassification errors due to sensor failures, occlusions, false negatives in CV models, etc. in the current perception modules. It should be noted that this proposition does not demand radical change to any of the current CV-based perception systems. Rather, KEP would complement the CV models by having the ability to leverage a machine-readable, unified knowledge representation in scene KG that is central to the KEP process.\\

\begin{tcolorbox}[title= Connection to Perception in Psychology]
Event perception is a particularly interesting area of research in Psychology where it is considered that ``the perception of events depends on both sensory cues and knowledge structures that represent previously learned information about event parts and inferences about actors’ goals and plans''\cite{zacks2007event}. The event segmentation theory (EST), a theory on the mechanism that segments ongoing activities into meaningful events, starts with a \textit{perceptual processing stream} that takes a representation of the current state of the world (e.g., perceptual information, language, external knowledge, etc.) and \textbf{produces a set of predictions} about ``what will happen a short time in the future'' \cite{zacks2020event}. This can be seen as an \textit{ongoing} comprehension process where the predictive processing is modulated by an (event) model that considers perceptual streams together with a representation of the world.\\ 

\textbf{\textit{Corollary to KEP}:} At a higher level, KEP functions on the same key fundamentals on which EST is built. First, they both start with a given set of percepts (or perceptual stream) as the input. Then, they rely on a rich representation of the world that is comprised of higher-level concepts and relations, similar to scene KG used by the KEP process. Given the perceptual stream and the representation of the world, EST predicts \textit{events} that will happen in a short time while KEP predicts both \textit{objects and events} that could be missing from the scene or may be seen within a short time. Viewed in this manner -- i.e. as an ongoing comprehension process of event perception -- KEP can be seen as an implementation of EST for improving machine perception in autonomous systems.

\end{tcolorbox}

\subsection{Other Benefits of KEP}
In addition to improving machine perception, the proposed KEP approach and the use of scene KG will have several other benefits. We briefly discuss them here.\\

\subsubsection{\textbf{External Knowledge Integration}}
Currently, all processes from perception to control primarily rely on large, diverse, and high-quality training data to function robustly. Despite the availability of a variety of useful knowledge sources, it remains a challenge to leverage them within current deep learning and computer vision methods. Interestingly, for KEP, the scene knowledge is represented in a KG and conformant to an ontology. This allows for easy integration of additional information, including geo-spatial data (e.g., Open Street Maps (OSM)), commonsense knowledge (e.g., ConceptNet, CSKG\cite{filip2021cskg}), driving rules and law (e.g., from driving manuals), etc. With this information represented as a KG, many existing knowledge-infused learning \cite{shadesof,valiant2006knowledge, garcez2020neurosymbolic} algorithms -- e.g., Knowledge Graph Embeddings (KGEs)\cite{rossi2021knowledge}, Graph Neural Networks (GNNs)\cite{wu2020comprehensive} -- can learn latent representations of this knowledge to be used with several downstream applications. Therefore, a scene KG continually evolving with KEP would create a rich internal representation of the external world (i.e. "world model") for autonomous systems. Recent work in the autonomous driving domain has shed some light into the development of KGs from real AD data -- e.g. the Driving Scene Knowledge Graph (DSKG) and enhancing it with geo-spatial data \cite{wickramarachchi2021knowledge} and commonsense knowledge \cite{chowdhury2021towards}.\\

\subsubsection{\textbf{Interpretability and Explainability}} Another benefit of KEP is the ability to use relational and contextual knowledge to generate explanations for the predictions. Consider the following example: Similar to the way a flight recorder (i.e. a black box) is used to trace the steps in-flight crash investigation, AD accident investigations look at replays of sensor readings, detected entities, and decisions made for each time point. Now, due to having semantic concepts, relations defined between entities, and raw sensor data mapped to their corresponding semantic types in the scene KG, it will allow investigators to interpret the raw sensor reading. The sensor data at the time of the crash can be easily interpreted as (i) \textit{high-speed} by comparing the recorded speed with the legal speed limit of the location, (ii) the applicable traffic rules can be obtained by first decoding the latitude/longitude readings as the higher-level concept \textit{``School Zone''} and then querying the rules applicable to school zones. Through such rich and contextual interpretations, we believe KEP could significantly improve the domain explainability of autonomous systems.

\begin{figure*}[ht]
    \centering
    \includegraphics[width=\textwidth]{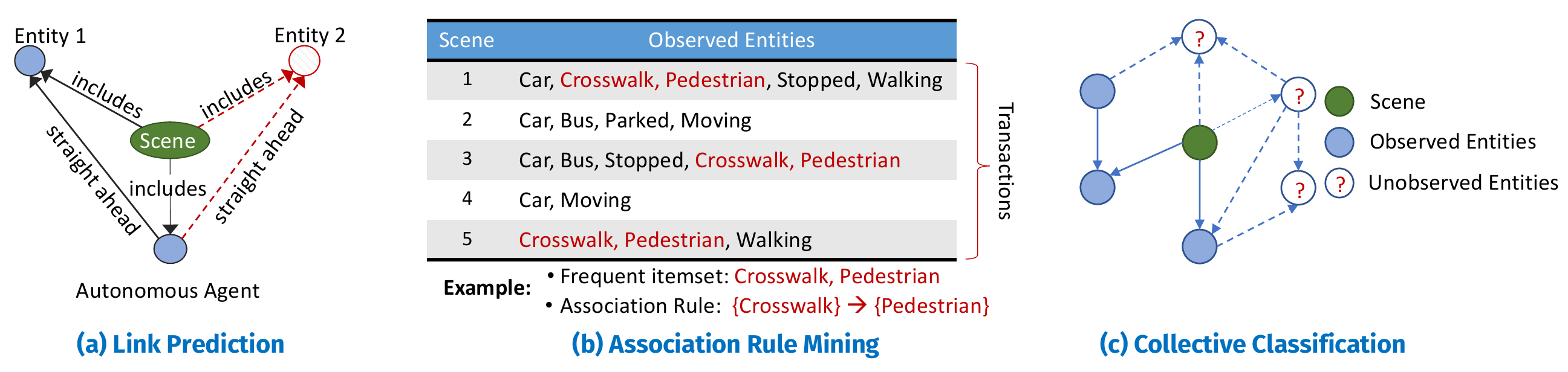}
    \caption{Three possible solutions devised for KEP based on different machine learning/data mining techniques}
    \label{fig:kep_solutions}
\end{figure*}

\section{The Formalization and Solutions for KEP}
\label{sec:formalization}

Next, we formalize KEP task as a generic knowledge completion problem. Then we devise possible solutions for KEP considering different machine learning and data mining techniques. 

\subsection{Formalization of KEP}
First, we define the notation for a formalization of KEP that would be generic enough to allow for a wide range of possible solutions. Let the KG $G: N\times R \times N$ where $N, R$ denote nodes and relations, respectively. Considering the different types of nodes in $G$, let $s^{(i)} \in S \subseteq N$ (where $S$ represents the set of scene nodes), $e_j \in E \subseteq N$ (where $E$ represents set of entity type nodes), and $\{E^{(i)}_{obs}, E^{(i)}_{inf}\} \subseteq  E \subseteq N$ (where $E^{(i)}_{obs}$ represents the subset of $E$ that are observed and included in $s^{(i)}$, and $E^{(i)}_{inf}$ as the subset of $E$ to be inferred). Scenes and entities are connected in $G$ through $(s^{(i)}, e_j) \in I \subseteq R$ where $I$ denotes the set of \texttt{includes} relations. Entities are connected with their types in $G$ through $T \subseteq R$ where $T$ denotes the set of \texttt{type} relations  -- e.g., \texttt{rdf:type}. Now, for a given scene $s^{(i)}$, the objective of KEP is to find  $E^{(i)}_{inf}$ such that  $E^{(i)}_{inf}  \Leftarrow \phi (G, s^{(i)}, E^{(i)}_{obs})$ where $\phi$ is the learned inference model generated and used by the solution.

\subsection{Possible Solutions for KEP}
\subsubsection{\textbf{Link Prediction}}

In the KG literature, Link Prediction (LP) is commonly used to address the KG incompleteness issue. While there are many techniques for LP, Knowledge Graph Embedding (KGE) methods, which represent KG nodes and relations in low-dimensional latent space, achieve the state-of-the-art performance for this task. Specifically, given an incomplete triple $(h, r, ?)$, KGE-based LP methods attempt to predict the missing component ($?$) by querying the KGE model $\phi(h, r, ?)$ to obtain the tail ($t$) of this triple. Given a KG with scenes and related entities, it begs the question of whether KEP can be formulated as a LP problem. In a scene KG, however, if $h=s^{(i)}$, predicting $t$ would mean predicting an entity instance (i.e. a specific entity) rather than an entity class (i.e. an entity type). One potential solution would be to reify the path from $s^{(i)}$ to the associated class of an entity included in the scene (e.g., \texttt{Car} included in a scene)\cite{wickramarachchi2021knowledge}. This path could be reified using a direct \texttt{\textbf{includesType}} relation as shown in Eq. 1. Note that this would allow the re-use of any existing KGE technique for predicting the tail of an \texttt{includesType} relation, thus predicting the entity type included in a scene.

\vspace{-1em}
\begin{gather*}
\small
    \langle s^{(i)}, \texttt{includes}, \texttt{e}_j \rangle \wedge
    \langle \texttt{e}_j, \texttt{rdfs:type}, \textcolor{red}{\texttt{?}} \rangle 
    \Rightarrow \\
    \langle s^{(i)}, \texttt{\textbf{includesType}}, \textcolor{red}{\texttt{?}} \rangle
\end{gather*}

\subsubsection{\textbf{Association Rule Mining}}
Association rule mining (ARM) is a popular data mining technique that is widely used to uncover interesting correlations, frequent patterns, and associations among sets of items. For example, in retail market basket analysis, ARM could be used to find associations between items customers frequently buy together. The objective of ARM is to generate \textit{association rules} in the form  $r_i: \{A, B\} \Rightarrow {\{C\}} | c$ where $\{A,B\}$ (i.e. antecedents) imply the co-occurrence of $\{C\}$ (i.e. consequent) (here, $ \{A, B\},  \{C\} \subset I$, referred to as \textit{itemsets}) with a confidence factor of $c;  0 \leq c \leq 1$. This would mean, at the minimum, $c\%$ transactions in the set of transactions $T$ satisfies the $r_i$ rule. Considering the objective of KEP, we may postulate whether the co-occurrences of entities could help predict the missing set of entities in a scene. If so, then KEP could be formulated as a market basket analysis problem, where the sets of items represent the set of observed entities in scenes. As shown in \cite{wickramarachchi2021knowledge}, the ARM approach for KEP consists of 3 main steps. First, an ARM algorithm (e.g., Apriori algorithm\cite{agrawalMSTV96}) is run on the training dataset (i.e. a scene KG) to generate a set of association rules. Second, a mask is created to filter the relevant set of rules for a given scene. A rule is considered relevant if its antecedent is a subset of the observed entities of the scene  (i.e. $E^{(i)}_{obs}$). Finally, the missing set of entities of this scene (i.e. $E^{(i)}_{inf}$) is obtained by considering the aggregation of consequents that satisfy the mask created above.\\ 

\subsubsection{\textbf{Collective Classification}}

\begin{figure*}[ht]
    \centering
    \includegraphics[width=0.9\textwidth]{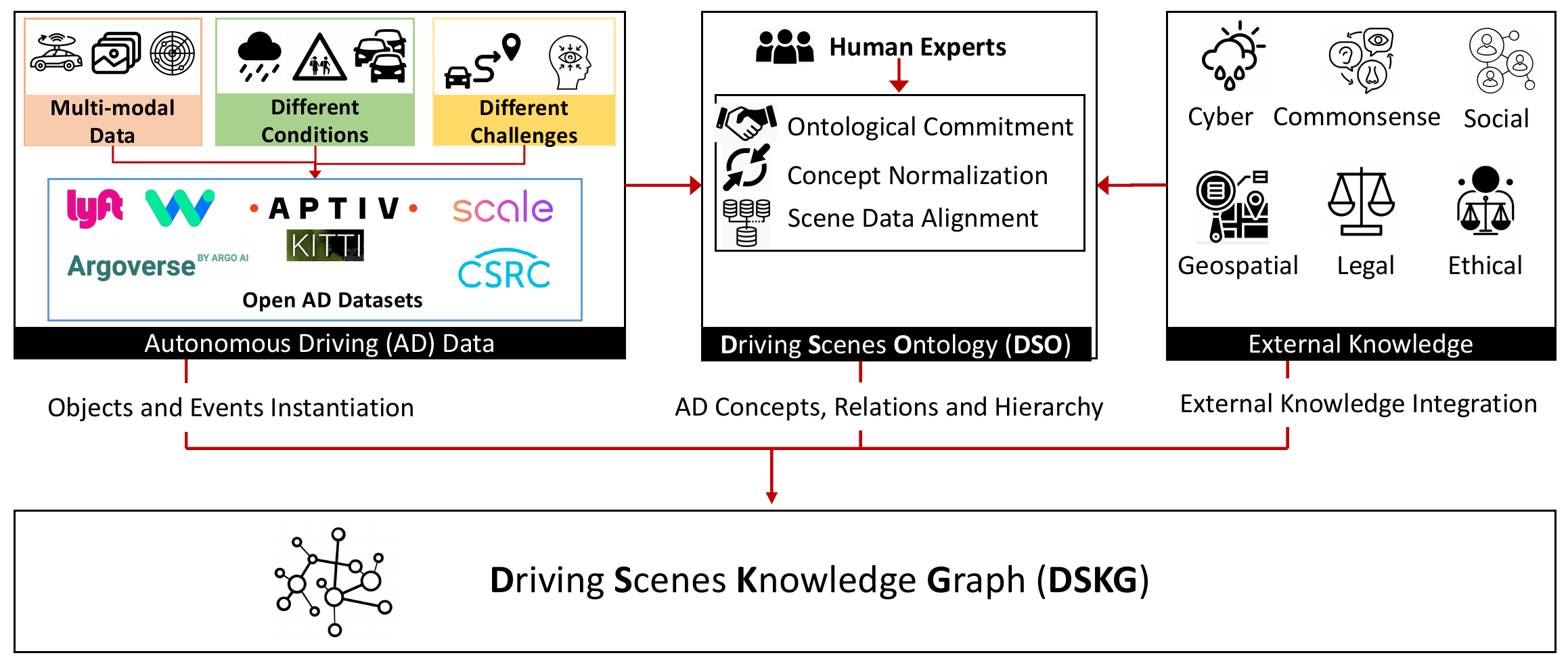}
    \caption{The process of creating a unified knowledge representation for knowledge-based downstream applications in AD domain}
    \label{fig:DSKG}
\end{figure*}

Collective classification is a combinatorial optimization problem that works on the premise that knowledge of the correct label for one node in a graph (i.e. semantic type in scene KG) improves the ability to assign correct labels to the other unlabeled nodes it connects to. In its most generic form, the goal of collective classification is to jointly determine the correct label assignments of all the nodes in a graph/network. Let $V={v_1, v_2,...,v_n}$ be the set of nodes in the graph, and let $N$ be the neighborhood function ($N \subseteq V$) that describes the underlying network structure\cite{Sen2010}. Each node in  $V$ is a random variable that can take a value from an appropriate domain, $L, L={l_1, l_2,...l_m}$. Considering the node label space, $V$ is divided into two sets: $X$, the nodes for which the correct label assignments are known (observed variables), and $Y$, the nodes whose values need to be inferred (unobserved variables). The inference task is to then label the nodes  $y_j \in Y$ with a subset of predefined labels in $L$. When considering KEP, nodes $X^{(i)}$ would denote the set of observed entities (i.e. $E^{(i)}_{obs}$) for scene $s^{(i)}$, while $y_j \in Y$ would denote the additional set of entities to be predicted/inferred (i.e. $E^{(i)}_{inf}$) where $E=L$ in the collective classification formalization.

\section{Case Studies}
\label{sec:case-studies}

In this section, we demonstrate the applicability of KEP in two application domains of autonomous systems: autonomous driving and smart manufacturing. We will also highlight the different types of knowledge KEP exploits in each use case.

\subsection{Autonomous Driving}
Autonomous driving is a key technical domain of interest in AI with the objective of achieving full (level-5) autonomy where the car drives itself without any human involvement. Key obstacles for achieving full autonomy center around the incompleteness in learning, dealing with edge-cases, and challenges associated with open-world driving environments. Currently, the computer vision based models solely rely on training from observational data, and there is no clear way to \textit{infuse} the innate understanding or knowledge that humans have about the world, including knowledge of different aspects of a scene, such as time, space, causality, semantics, etc. We believe one possible way of tackling this problem could be to change the way the scenes are represented in AD. We argue that AD requires a more holistic representation of driving scenes composed of higher-level semantic concepts and relations, that evolves over time, and contains high-quality multi-modal data from heterogeneous sources. Knowledge Graphs (KGs) naturally fit these requirements as they are capable of representing expressive and meaningful relations among entities (i.e. objects, events, and abstract concepts) in the real world.

Figure \ref{fig:DSKG} shows a schematic overview of how the scene KG is developed as a unified representation for AD knowledge. KEP, through a \textit{cyclic process},  updates this knowledge over time by predicting the unseen entities leveraging the underlying relational structure and diverse knowledge in the scene KG. We believe this would significantly improve the machine perception in AD (i.e. by reducing misclassification errors, omissions, object occlusions, etc.). Additionally, the latent transformations of the scene KG, such as knowledge graph embeddings\cite{wickramarachchi2020evaluation}, would allow the subsequent processes in AD to benefit from the relational knowledge in the scene KG. This could push the current technology one step closer to achieving full autonomy.

\subsection{Smart Manufacturing}
Smart manufacturing (or Industry 4.0/I4.0) \cite{wang2021smart} employs smart robots that function fully autonomously for a defined task. Precise understanding of manufacturing events plays a critical role in such systems. It requires analyzing an array of sensor feeds (e.g.,  video streams from various cameras, pressure, thermal, and humidity sensors, etc.) as well as understanding the ``manufacturing context” through the use of other sensor data, external knowledge and information relevant to the production line. Similar to the way driving scenes are represented above for autonomous driving, we argue the manufacturing process can be seen as a stream of ``manufacturing scenes”. In this case, a KG can be developed to represent manufacturing scenes that are composed of entities (i.e. objects and events) along with their spatial and temporal attributes and relevant meta-data. In addition to scene data, the KG allows for the seamless integration of existing knowledge sources such as ontologies for I4.0 (e.g., Standards Ontology, Resource Reconfiguration Ontology, Manufacturing Resource Capability Ontology
(MaRCO), I4.0 Components Ontology, etc.)\cite{yahya2021industry} as well as relevant domain knowledge (e.g., data from robot product specifications, safe working conditions, etc.). In this context, the KEP task could be used for improving event understanding by predicting the possible events of failures and/or outliers in components (e.g., prognostic component)  of smart manufacturing systems.

\section{Experimental Results}
\label{sec:results}

\begin{table*}[ht]
\centering
\caption{KEP results of Pandaset (DSKG-P$_R$)\cite{wickramarachchi2021knowledge} on 3 algorithms, each experiment averaged with standard deviation across 5 runs. Evaluation metrics: MRR=Mean Reciprocal Rank, H@K= Hits@K, Accu. = KEP Accuracy, Micro/Macro F1 = Micro/Macro-averaged-F1-score}
\label{tab:kep-results}
\begin{tabular}{l|cccc|ccc} 
\toprule
       & \multicolumn{4}{c|}{Ranking Metrics}                                              & \multicolumn{3}{c}{KEP Performance Metrics}                    \\ 
\cline{2-8}
       & MRR                & H@1                & H@3                & H@10               & Accu. (\%)          & Micro F1           & Macro F1            \\ 
\hline
TransE\cite{bordes2013translating} & 0.32 \textpm  0.03          & 0.16 \textpm 0.05          & 0.35\textpm  0.04          & 0.71\textpm  0.03          & 22.98 \textpm 4.33          & 0.26 \textpm 0.04          & 0.20 \textpm 0.02           \\
HolE\cite{nickel2016holographic}   & \textbf{0.93 \textpm 0.00} & \textbf{0.87\textpm  0.01} & \textbf{0.98 \textpm 0.00} & \textbf{1.00\textpm  0.00} & \textbf{88.91\textpm  0.64} & \textbf{0.90 \textpm 0.01} & \textbf{0.87 \textpm 0.00}  \\
ConvKB\cite{dai2018novel} & 0.29\textpm  0.01          & 0.11 \textpm 0.02          & 0.31 \textpm 0.02          & 0.86\textpm  0.02          & 17.83\textpm  1.99          & 0.22\textpm  0.02          & 0.17 \textpm 0.02           \\
\bottomrule
\end{tabular}
\end{table*}

To provide a synopsis of how KEP performs on real-world autonomous systems, we present experimental results of an LP-based implementation of KEP considering autonomous driving data. For this purpose, we refer to the Driving Scenes Knowledge Graph (DSKG) introduced by Wickramarachchi et al.\cite{wickramarachchi2021knowledge} which represents the scene data from the Pandaset\cite {pandaset} dataset and is conformant to the Driving Scenes Ontology (DSO). Table \ref{tab:kep-results} summarizes the KEP results with three KGE algorithms and two sets of evaluation metrics -- i.e. ranking metrics (hits@K, Mean Reciprocal Rank (MRR)) and KEP performance metrics (Accuracy, Micro/Macro F1). This evaluation supports the hypothesis for KEP, predicting missing entities with a 0.87 precision (hits@1) and 88.91\% accuracy. A more detailed and thorough evaluation with other AD datasets, different KG structures, and other implementations can be found in \cite{wickramarachchi2021knowledge}. When considering the space and time complexities of this solution, it is most efficient when TransE is used (Time: $\mathcal{O}(n_t d)$, Space: $\mathcal{O} (m d + n d)$) and least efficient when ConvKB is used (Space: $\mathcal{O} (m d + n d + (\tau +3)d)$, Time: $\mathcal{O}(n_t \tau d)$). It should be noted, however, the space/time complexities of KEP are dependent on the complexity of the underlying solution and its sub-components.

\section {Conclusions and Future work}
\label{sec:conclusion}

This paper introduces the novel task of knowledge-based entity prediction (KEP) that leverages knowledge from heterogeneous sources for inferring and/or predicting unrecognized entities. A new proposition has been presented for using KEP within a \textit{cyclic perception process} for improving machine perception in autonomous systems. We define and formalize KEP as a knowledge completion problem and present three possible solutions considering several data mining and machine learning techniques. Finally, we demonstrate the applicability of KEP for two use-cases of autonomous systems from different domains. In the near future, we plan to implement and evaluate the solutions discussed in this paper with real autonomous systems data. More generally, however, we hope the proposed reformulation of KEP may inspire other researchers to devise, apply, and share new and novel solutions to the KEP problem.

\section*{Acknowledgements}
We thank Dr. Valerie Shalin for the helpful suggestions and discussion on perception in Psychology. This work was supported in part by Bosch Research, NSF grants \#2133842, and \#2119654. Any opinions, findings, and conclusions or recommendations expressed in this material are those of the authors and do not necessarily reflect the views of the funding organizations. We thank the anonymous reviewers for their constructive comments.\\

\bibliographystyle{IEEEtran}
\bibliography{references.bib}

\begin{thebibliography}{10}
\providecommand{\url}[1]{#1}
\csname url@samestyle\endcsname
\providecommand{\newblock}{\relax}
\providecommand{\bibinfo}[2]{#2}
\providecommand{\BIBentrySTDinterwordspacing}{\spaceskip=0pt\relax}
\providecommand{\BIBentryALTinterwordstretchfactor}{4}
\providecommand{\BIBentryALTinterwordspacing}{\spaceskip=\fontdimen2\font plus
\BIBentryALTinterwordstretchfactor\fontdimen3\font minus
  \fontdimen4\font\relax}
\providecommand{\BIBforeignlanguage}[2]{{%
\expandafter\ifx\csname l@#1\endcsname\relax
\typeout{** WARNING: IEEEtran.bst: No hyphenation pattern has been}%
\typeout{** loaded for the language `#1'. Using the pattern for}%
\typeout{** the default language instead.}%
\else
\language=\csname l@#1\endcsname
\fi
#2}}
\providecommand{\BIBdecl}{\relax}
\BIBdecl

\bibitem{ramanishka2018toward}
V.~Ramanishka, Y.-T. Chen, T.~Misu, and K.~Saenko, ``Toward driving scene
  understanding: A dataset for learning driver behavior and causal reasoning,''
  in \emph{Proceedings of the IEEE Conference on Computer Vision and Pattern
  Recognition}, 2018, pp. 7699--7707.

\bibitem{grigorescu2020survey}
S.~Grigorescu, B.~Trasnea, T.~Cocias, and G.~Macesanu, ``A survey of deep
  learning techniques for autonomous driving,'' \emph{Journal of Field
  Robotics}, vol.~37, no.~3, pp. 362--386, 2020.

\bibitem{zacks2007event}
J.~M. Zacks, N.~K. Speer, K.~M. Swallow, T.~S. Braver, and J.~R. Reynolds,
  ``Event perception: a mind-brain perspective.'' \emph{Psychological
  bulletin}, vol. 133, no.~2, p. 273, 2007.

\bibitem{zacks2020event}
J.~M. Zacks, ``Event perception and memory,'' \emph{Annual Review of
  Psychology}, vol.~71, pp. 165--191, 2020.

\bibitem{filip2021cskg}
F.~Ilievski, P.~Szekely, and B.~Zhang, ``Cskg: The commonsense knowledge
  graph,'' in \emph{The Semantic Web}, R.~Verborgh, K.~Hose, H.~Paulheim, P.-A.
  Champin, M.~Maleshkova, O.~Corcho, P.~Ristoski, and M.~Alam, Eds.\hskip 1em
  plus 0.5em minus 0.4em\relax Cham: Springer International Publishing, 2021,
  pp. 680--696.

\bibitem{shadesof}
A.~{Sheth}, M.~{Gaur}, U.~{Kursuncu}, and R.~{Wickramarachchi}, ``Shades of
  knowledge-infused learning for enhancing deep learning,'' \emph{IEEE Internet
  Computing}, vol.~23, no.~6, pp. 54--63, Nov 2019.

\bibitem{valiant2006knowledge}
L.~G. Valiant, ``Knowledge infusion,'' 2006.

\bibitem{garcez2020neurosymbolic}
A.~d. Garcez and L.~C. Lamb, ``Neurosymbolic ai: The 3rd wave,'' \emph{arXiv
  preprint arXiv:2012.05876}, 2020.

\bibitem{rossi2021knowledge}
A.~Rossi, D.~Barbosa, D.~Firmani \emph{et~al.}, ``Knowledge graph embedding for
  link prediction: A comparative analysis,'' \emph{ACM Trans. Knowl. Discov.
  Data}, vol.~15, no.~2, Jan. 2021.

\bibitem{wu2020comprehensive}
Z.~Wu, S.~Pan, F.~Chen, G.~Long, C.~Zhang, and S.~Y. Philip, ``A comprehensive
  survey on graph neural networks,'' \emph{IEEE transactions on neural networks
  and learning systems}, vol.~32, no.~1, pp. 4--24, 2020.

\bibitem{wickramarachchi2021knowledge}
R.~Wickramarachchi, C.~Henson, and A.~Sheth, ``{Knowledge-infused Learning for
  Entity Prediction in Driving Scenes},'' \emph{{Frontiers in Big Data}},
  vol.~4, p. 759110, 2021.

\bibitem{chowdhury2021towards}
S.~N. Chowdhury, R.~Wickramarachchi, M.~H. Gad-Elrab, D.~Stepanova, and
  C.~Henson, ``Towards leveraging commonsense knowledge for autonomous
  driving,'' in \emph{The 20th International Semantic Web Conference (ISWC)},
  2021.

\bibitem{agrawalMSTV96}
R.~Agrawal, H.~Mannila, R.~Srikant, H.~Toivonen, and A.~I. Verkamo, ``Fast
  discovery of association rules,'' in \emph{Advances in Knowledge Discovery
  and Data Mining}, U.~M. Fayyad, G.~Piatetsky{-}Shapiro, P.~Smyth, and
  R.~Uthurusamy, Eds.\hskip 1em plus 0.5em minus 0.4em\relax {AAAI/MIT} Press,
  1996, pp. 307--328.

\bibitem{Sen2010}
P.~Sen, G.~Namata, M.~Bilgic, and L.~Getoor, \emph{Collective
  Classification}.\hskip 1em plus 0.5em minus 0.4em\relax Boston, MA: Springer
  US, 2010, pp. 189--193.

\bibitem{wickramarachchi2020evaluation}
R.~Wickramarachchi, C.~Henson, and A.~Sheth, ``{An evaluation of knowledge
  graph embeddings for autonomous driving data: Experience and practice},''
  \emph{AAAI 2020 Spring Symposium on Combining Machine Learning and Knowledge
  Engineering in Practice (AAAI-MAKE 2020)}, 2020.

\bibitem{wang2021smart}
B.~Wang, F.~Tao, X.~Fang, C.~Liu, Y.~Liu, and T.~Freiheit, ``Smart
  manufacturing and intelligent manufacturing: A comparative review,''
  \emph{Engineering}, vol.~7, no.~6, pp. 738--757, 2021.

\bibitem{yahya2021industry}
M.~Yahya, J.~G. Breslin, and M.~I. Ali, ``{Semantic web and knowledge graphs
  for industry 4.0},'' \emph{Applied Sciences (Switzerland)}, vol.~11, no.~11,
  pp. 1--23, 2021.

\bibitem{bordes2013translating}
A.~Bordes, N.~Usunier, A.~Garcia \emph{et~al.}, ``Translating embeddings for
  modeling multi-relational data,'' in \emph{Adv. in Neural Information
  Processing Systems}, 2013, pp. 2787--2795.

\bibitem{nickel2016holographic}
M.~Nickel, L.~Rosasco, and T.~Poggio, ``Holographic embeddings of knowledge
  graphs,'' in \emph{Thirtieth AAAI Conference on Artificial Intelligence},
  2016.

\bibitem{dai2018novel}
D.~Q. Nguyen, T.~D. Nguyen, D.~Q. Nguyen \emph{et~al.}, ``A novel embedding
  model for knowledge base completion based on convolutional neural network,''
  in \emph{Proceedings of NAACL-HLT}, 2018, pp. 327--333.

\bibitem{pandaset}
{Scale AI}, ``Pandaset open datasets - scale,''
  \url{https://scale.com/open-datasets/pandaset}, 2020, (Accessed on
  01/25/2021).

\end{thebibliography}

\section*{Authors}

\noindent \textbf{Ruwan Wickramarachchi:} Ruwan is a Ph.D. student at the AI Institute, University of South Carolina. His dissertation research focuses on introducing expressive knowledge representation and knowledge-infused learning techniques to improve machine perception and context understanding in autonomous systems (e.g., autonomous driving and smart manufacturing). Contact him at: \texttt{ruwan@email.sc.edu} \\

\noindent \textbf{Cory Henson:}
Cory is a lead research scientist at the Bosch Research and Technology Center in Pittsburgh, PA. His research focuses on the application of knowledge representation and knowledge-infused learning to enable autonomous driving. He also holds an Adjunct Faculty position at Wright State University. Prior to joining Bosch, he earned a Ph.D. in Computer Science from WSU, where he worked at the Kno.e.sis Center applying semantic technologies to represent and manage sensor data on the Web. Contact him at: \texttt{cory.henson@us.bosch.com} \\

\noindent \textbf{Amit Sheth:} Amit is the founding director of the AI Institute, University of South Carolina. His current core research is in knowledge-infused learning and explanation, and transnational research includes personalized and public health, social good, education, and future manufacturing. He is a fellow of IEEE, AAAI, AAAS, and ACM. Contact him at: \texttt{amit@sc.edu} (http://aiisc.ai/amit)\\

\end{document}